\UseRawInputEncoding
\documentclass[journal]{IEEEtran} 

\IEEEoverridecommandlockouts

\usepackage{moreverb,url}
\usepackage{caption}

\usepackage[colorlinks,bookmarksopen,bookmarksnumbered,citecolor=red,urlcolor=red]{hyperref}

\usepackage{algorithm}
\usepackage{algpseudocode}
\usepackage{cite}
\usepackage{amsmath,amssymb,amsfonts}
\usepackage{graphicx}
\usepackage{textcomp}
\usepackage[table, dvipsnames]{xcolor}
\usepackage{float}
\usepackage{subcaption}
\usepackage{tabularray}
\usepackage{multirow}
\usepackage{booktabs}
\usepackage{hyperref}
\usepackage{titlesec}
\usepackage{tikz}
\usetikzlibrary{arrows.meta,positioning,calc}
\usepackage[disable]{todonotes}
\DeclareCaptionFont{mysize}{\fontsize{8}{10}\selectfont}
\usepackage{soul}
\usepackage[subtle,tracking=normal]{savetrees} 
\usepackage{balance}
\usepackage{tikz}
\usetikzlibrary{shapes.geometric, arrows, positioning, fit, backgrounds}

\newcommand\BibTeX{{\rmfamily B\kern-.05em \textsc{i\kern-.025em b}\kern-.08em
T\kern-.1667em\lower.7ex\hbox{E}\kern-.125emX}}

\def\BibTeX{{\rm B\kern-.05em{\sc i\kern-.025em b}\kern-.08em
    T\kern-.1667em\lower.7ex\hbox{E}\kern-.125emX}}

\newcommand{\multilinecomment}[1]{}

\author{Andrea Dal Prete$^{1,2}$, Seyram Ofori$^{2}$, Chan Yon Sin$^{3}$, Ashwin Narayan$^{2}$, \\Ding Shuo$^{4}$, Francesco Braghin$^{1}$, Haoyong Yu$^{2}$, and Marta Gandolla$^{1}$ 
\thanks{$^{1}$ Department of Mechanical Engineering, Politecnico di Milano, Italy. $^{2}$ Department of Biomedical Engineering, National University of Singapore (NUS), Singapore. $^{3}$ School of Science and Technology, Singapore University of Social Sciences (SUSS), Singapore. $^{4}$ College of Mechanical and Electrical Engineering, Nanjing University of Aeronautics and Astronautics, Nanjing, China.}
\thanks{Corresponding author e-mail: {andrea.dalprete@polimi.it}.}}

\begin{document}


\title{Toward Context-Aware Exoskeleton Assistance: Integrating Computer Vision Payload Estimation with a Multi-Metric Optimization Space}

\maketitle

\begin{abstract}

Back-support exoskeletons mitigate musculoskeletal strain, yet current systems rely on reactive sensing and lack context-aware assistance modulation. This paper presents a population-derived optimization framework and a predictive vision-based adaptive control strategy. First, we construct a multi-metric optimization space combining electromyography reduction, perceived discomfort, and user preference, revealing a non-linear relationship between payload and optimal assistance from experiments with 12 subjects. Second, we develop a computer vision-based adaptive control leveraging a fine-tuned vision transformer (DINOv2) and depth sensing to estimate payloads prior to lifting, eliminating actuation latency. Validation with an additional 12 subjects demonstrates robust payload estimation (82.41\% accuracy). The proposed adaptive strategy reduces peak back muscle activation by up to 23\% and improves average offloading by 8.15\% over static baselines, without increasing discomfort. These results highlight the benefits of predictive perception and user-centric optimization for enhanced human-exoskeleton interaction.
\end{abstract}


\section{Introduction}
\subsubsection*{Current Limitations in Active Back-Support Exoskeletons}
Musculoskeletal disorders are a growing concern in industrial workplaces, particularly in construction, production, and logistics, where workers regularly lift heavy loads. To address these issues, back-support exoskeletons have been developed to reduce muscular activation and spinal loads, which are key contributors to back impairment \cite{Have, Dolan, NIOSH, LoozeBackPain}. Over the past few decades, design advancements have improved active back-support exoskeletons in terms of kinematic compatibility, weight, and acceptability \cite{Ding, Chung, Roveda, ToxiriRoboMate, BBEX}. 

Despite these improvements, challenges remain regarding optimization and adaptation; current back-support exoskeletons lack effective mechanisms to integrate user and environmental data for high-level control, leaving significant performance gains untapped \cite{DalPrete}. For instance, although maintaining high assistance regardless of payload may seem valuable for reducing muscular effort, studies indicate that benefits plateau at higher support levels \cite{Walter2023}. Furthermore, increasing support can cause joint misalignment, discomfort, and reduced acceptability, highlighting the need to balance muscular activity, perceived discomfort, and user preference \cite{Luger2023}, \cite{Ingraham}. Introducing assistance dependencies on payload adds further uncertainty regarding the best support selection. Thus, selecting optimal support levels based on environmental conditions or user intent remains an open research question. 

A contrasting but key challenge concerns the exoskeleton's perception. Adaptive control is feasible only in active exoskeletons, where real-time payload estimation and adaptive modulation are critical for enhancing performance \cite{DalPrete, Ali}. Notably, the highest back muscle activation occurs during the initial lifting phase \cite{vonArx2021}, emphasizing the need for rapid support modulation, but existing payload estimation techniques often fail to address this latency, limiting their effectiveness. 

\subsubsection*{Past research}
Previous researchers have proposed strategies to exploit adaptation for payload estimation and environmental comprehension. One of the first solutions by Mateos \cite{Mateos2017} employed force sensing resistor insoles to measure weight changes and distinguish lifting phases. Related research by Matijevich \textit{et al}. \cite{Matijevich} concluded that pressure insoles are essential for proper back load estimation. Lazzaroni \textit{et al}. \cite{Lazzaroni}, \cite{Lazzaroni2019} and Toxiri \textit{et al}. \cite{Toxiri2018} used optimized electromyography (EMG) armbands to modulate support based on grasping strength. While potentially significant, this strategy is limited by subject-specific behaviors, EMG drawbacks, and ergonomic considerations. Similarly, Islam \textit{et al}. \cite{Islam2019} and Tahir \textit{et al}. \cite{Tahir} proposed force-myography sensors, though these face challenges with sensor placement sensitivity and generalization \cite{Xiao}. Pesenti \textit{et al}. \cite{Pesenti2023} explored long short-term memory models using IMU kinematics, while others integrated natural language processing for user-driven modulation \cite{Olmo2024}. However, these approaches often suffer from latency and long inference times. As noted, since peak activation occurs at the lifting cycle's start, early-phase support is crucial. In this context, computer vision, which has been explored in other domains for locomotion mode recognition \cite{11122304, laschowski2021computer}, could represent a solution by enabling the exoskeleton to see the environment, and infer the payload from distance.

\subsubsection*{Proposed Approach and Contributions}
To address these gaps, this manuscript explores a novel optimization framework for active back-support exoskeletons, considering payload estimation, muscular activity reduction, user discomfort, and preferences. We then present a computer vision-based pipeline leveraging state-of-the-art object detection and payload estimation to enable operation in high-performance regions. Specifically, the contributions include:
\begin{itemize}
\item \textbf{Optimization space definition.} We formulate a novel optimization space defined by payload and assistance level, conducting a baseline campaign with 12 subjects. Using this data, we develop a smooth performance optimization function based on EMG reduction, discomfort, and preference to identify performance-maximizing regions.
\item \textbf{Vision-based payload estimation pipeline.} Building on the optimization analysis, we design a real-time computer vision pipeline to enhance environmental comprehension. The pipeline integrates deep learning models to detect objects, predict payloads, and adapt support levels according to the optimal trends found.
\item \textbf{Integration and validation.} We integrate the vision-based pipeline into the exoskeleton control loop and validate its performance with 12 subjects, assessing EMG signals, perceived discomfort, and user preference.
\item \textbf{Open-source contribution.} We provide open-source data and code as Supplementary Material to facilitate further scientific exploration.
\end{itemize}
This study tackles two key challenges: defining optimal support regions and enabling rapid, subject-independent payload estimation. By addressing these, we aim to establish new benchmarks for active back-support exoskeletons in industrial settings while enhancing contextual awareness.

\emph{Outline.} Section~\ref{sec:Methods} describes the optimization formulation, environmental pipeline, and hardware. Section~\ref{sec:Experiments} details baseline and validation experiments. Results and discussion are presented in Section~\ref{sec:Results} and Section~\ref{sec:Discussion}, followed by conclusions in Section~\ref{sec:Conclusion}.

\section{Methods}
\label{sec:Methods}
\subsection{Exoskeleton Optimization Searching Space}
\label{sec:ExoOptSpace}

\begin{figure*}[t!]
\centering
\includegraphics[width=1\linewidth]{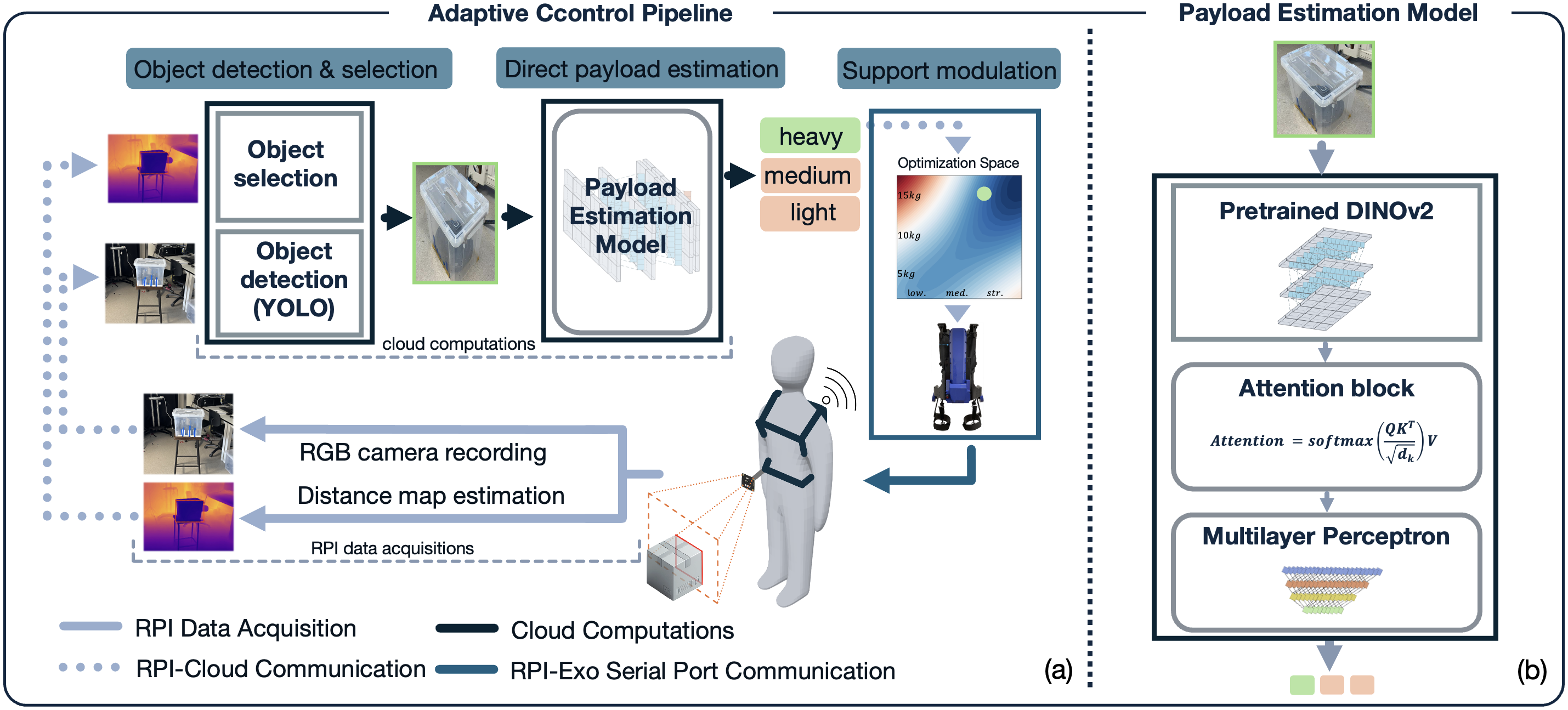}
\caption{Overview of the computer vision-based adaptive control} pipeline (a): images captured by the Raspberry Pi 4B (RPI) are transmitted to the cloud workstation for object detection and payload estimation, and the resulting predictions are returned to the RPI for serial communication with the exoskeleton. Panel (b) illustrates the architecture of the DINOv2-based payload-estimation model.
\label{fig:Pipeline&PylEstModels}
\end{figure*}

As introduced, in the first part of our research, we aim to identify optimal operating regions for back-support exoskeletons using two variables: assistance level ($A_s$) and user payload ($P_{yl}$). $A_s$ is a controllable parameter, while $P_{yl}$ is determined by the environment. In Section \ref{subsec:BaselineResults}, we illustrate the resulting Optimization Functions ($J_j$) within this space, highlighting optimal operating regions. To formulate these functions, we sampled discrete points with 12 subjects (see Section~\ref{subsec:BaselineExperiments}) to generate a dataset:
\begin{equation}\label{eq:Equation1}
    D = \{x^i,t^i\}^N_i
\end{equation}
where $x^i = (A_s^i, P_{yl}^i)$ are assistance-payload pairs, $t^i$ are objective function samples ($J_j^i$), and $N$ represents the number of samples in the dataset. Using Gaussian processes with a radial basis function kernel \cite{GPy, Rasmussen2006Gaussian}, we fit the dataset $D$. We optimized the Gaussian processes model hyperparameters (kernel length-scales and observation noise) by maximizing the marginal log-likelihood of the training data. This Bayesian regression approach allows us to construct a continuous Optimization Functions (OF):
\begin{equation}\label{eq:Equation2}
J_j(A_s, P_{yl}) = t, \quad J_j: \mathbb{R}^2 \rightarrow \mathbb{R}.
\end{equation}
The Total Optimization Functions (total-OF) combines three components: i) EMG activity reduction ($J_{EMG}$), ii) perceived discomfort ($J_{dsc}$), and iii) user preference ($J_{prf}$). $J_{EMG}$ measures the percentage reduction in muscular activity compared to the no-exoskeleton condition. For $J_{dsc}$, we assess discomfort with a questionnaire with nine weighted questions ($q_i$):
\begin{enumerate}
\item I experience uncomfortable forces compared to other conditions.
\item I feel highly comfortable in the current setting. $\textcircled{-}$
\item I had to work against the assistance in the current setting.
\item The exoskeleton restricts my natural movement in the current setting.
\item I have to think more about my movements in the current setting.
\item I felt nervous and unsafe in the current setting.
\item My fatigue is reduced in the current setting. $\textcircled{-}$
\item I am satisfied with the current setting. $\textcircled{-}$
\item The exoskeleton is useful for my task in the current setting. $\textcircled{-}$
\end{enumerate}
Questions are rated from 1 (strongly disagree) to 5 (strongly agree). Weights prioritize factors directly related to discomfort ($w_{1,2,4}=2$, $w_3=1.5$) over indirect connections ($w_{5-9}=1$). Scores are flipped for positively framed questions (2, 7, 8, 9, highlighted with a $\textcircled{-}$ sign). The final discomfort sample is:
\begin{equation}\label{eq:Equation3}
J_{dsc}(A_s^i, P_{yl}^i) = \sum_{k=1}^{9} w_k q^i_k.
\end{equation}
For $J_{prf}$, subjects indicate their preferred assistance (light vs. strong) when lifting light ($5$ kg) and heavy ($15$ kg) weights. After interpolation and normalization, the total function is:
\begin{equation}\label{eq:Equation4}
J_{tot}(A_s, P_{yl}) = 0.6 J_{EMG} - 0.2 J_{dsc} + 0.2 J_{prf}.
\end{equation}
We weight $J_{EMG}$ highest as it provides a quantitative measure, while we weight qualitative metrics ($J_{dsc}, J_{prf}$) lower. The total function $J_{tot}$ reflects the negative cost of operating in a region, with higher values indicating better performance, guiding the selection of optimal assistance levels.
\subsection{High-Level Control Strategy for Back Exoskeleton Optimization}\label{subsec:HL_Control_Strategy_ForBE}
Following the optimization formulation in Section~\ref{sec:ExoOptSpace}, our results indicate that static support fails to maximize performance across varying payloads. Consequently, the back-support exoskeleton requires payload-adaptive control. To achieve this, we developed the computer vision-based adaptive control pipeline (Fig.~\ref{fig:Pipeline&PylEstModels}-a). This pipeline integrates RGB and depth data to detect and select the target object. The resulting image is processed by a fine-tuned vision transformer \cite{Dosovitskiy2021} to classify the weight (light, medium, or heavy), determining the optimal assistance level. This approach specifically addresses scenarios where object weight is unknown (e.g., transparent boxes), requiring inference based on visual content. The pipeline components are detailed below.
\subsubsection{Object Detection \& Selection}\label{subsec:Object Detection&Selection}
The first stage of the pipeline performs object detection and selection. To enhance environmental awareness, we fine-tuned YOLOv11 \cite{Redmon} on a custom dataset of 831 images collected at the BioRobotics Lab at the National University of Singapore, labeled with a ``box'' class and bounding box coordinates, and split into 88\% training, 9\% validation, and 3\% testing. The fine-tuned model runs on cloud GPUs in real-time, processing RGB frames from the embedded camera. Since YOLO may detect multiple objects, a Time-of-Flight (ToF) distance sensor provides approximate depth estimates, which we use to identify the object most likely to be picked up. Although we tested state-of-the-art monocular depth estimation models such as DepthAnythingv2 \cite{Yang2} and MiDaS \cite{Ranftl} fine-tuned for metric depth estimation (see Supplementary Materials), the Arducam ToF camera offered superior performance indoors within our operational range ($[0\!-\!4]\,\mathrm{m}$). Using the distance map, we compute a probability factor $\alpha_i$ for each detected object:
\begin{equation}
\alpha_i = 10 \cdot e^{-\lambda_{\theta}|\theta_i|} \cdot e^{-\lambda_{d}d_i}
\end{equation}
where $\lambda_{\theta}=0.1$ and $\lambda_d=0.5$ are the angular and distance decay coefficients, respectively (optimized during preliminary tests), $\theta_i$ is the angle of the $i$-th object in the camera’s field of view (with $0$ at the center), $d_i$ is the object’s distance, and 10 is a scale factor. We assign lower probabilities to objects farther from the user or near the field boundaries. After softmax normalization, we select the object with the highest pick-up probability. If this object is within $2\,\mathrm{m}$, we consider the lifting cycle imminent, block the recognition pipeline, crop the environment image to its bounding box, and pass it to the next layer.
\subsubsection{Payload Estimation}\label{subsec:PylEst}
After the object detection and selection layer (Section~\ref{subsec:Object Detection&Selection}), we feed the cropped image of the selected object into a fine-tuned deep-learning model for direct weight classification (light, medium, heavy). Rather than relying solely on single-object detection, our pipeline introduces an additional inference layer to enhance contextual understanding, enabling robust estimation even in cluttered or ambiguous scenarios (e.g., multiple items within a box). To this end, we compare two state-of-the-art architectures: MobileNetv2 \cite{Sandler}, a Convolutional Neural Network (CNN) optimized for real-time performance, and DINOv2, a vision transformer-based foundation model known for high-quality feature extraction, contextual understanding, and strong transfer learning capabilities \cite{Yang, ZhangObj, Oquab, Cui, Dermyer}.

MobileNetv2 Approach:
MobileNetv2 processes input images of size $(256,256,3)$ and outputs feature maps of shape $(1,8,8,1024)$, which we flatten into 65{,}536 features. We project these into a 32-dimensional latent space through a four-layer fully connected network. To integrate physical information, we concatenate bounding box dimensions and object distance with the extracted vector, yielding a 35-dimensional input to a final linear classifier. We use ReLU activations, softmax output, Root Mean Square Propagation (RMSprop) optimizer ($lr=1e^{-4}$), and categorical cross-entropy loss. In this configuration, MobileNetv2 serves as a frozen feature extractor, and we train only the subsequent layers for classification on the latent features extracted.

DINOv2 Approach (Fig.~\ref{fig:Pipeline&PylEstModels}-b):
We use the pretrained DINOv2 base model \cite{PatrickLabatut} and append a single-head attention block. We average the output features of shape $(1,324,768)$ into a 768-dimensional vector and pass it to a multilayer perceptron with dropout (0.4), ReLU activation, and a softmax classifier. Since DINOv2 has demonstrated strong depth-perception capabilities in prior work, we do not add any additional physical features. We use the Adam optimizer for training ($lr=1e^{-4}$) with categorical cross-entropy loss. Here, we fine-tune DINOv2 by freezing its core weights and updating only the appended layers.

We trained both models on a dataset of 5480 labeled images (32.03\%, 34.01\%, and 33.96\% for light, medium, and heavy), collected at the BioRobotics Lab at the National University of Singapore. To ensure robust generalization, the dataset contains the same objects and translucent boxes used in the validation studies, supplemented with various non-transparent boxes to increase visual variance. Furthermore, we systematically injected environmental variability by capturing images from multiple viewing angles, under diverse lighting conditions (e.g., natural sunlight, artificial room lighting, and surface reflections), and across different spatial contexts, ranging from clean floors and tables to heavily cluttered backgrounds. We trained the MobileNetv2-based classifier for 40 epochs with a batch size of 400, whereas we fine-tuned DINOv2 for 20 epochs with a batch size of 40. We provide codes, datasets, and additional details on hyperparameter choices in the Supplementary Materials.

\subsubsection{Support Modulation}
To modulate the exoskeleton support between low, medium, and strong, we use a simple parameter $k_{pyl}$ to scale the low-level assistive torque profile provided to the hip joints within a lifting cycle, described in Section \ref{subs:low_level_contorl}. We find subject-specific values for $k_{pyl}^{min}$, $k_{pyl}^{max}$ during experiments as explained in section~\ref{subsec:BaselineExperiments} to find the minimum and maximum peak value of assistance which the subject feels comfortable with while lifting a light weight ($5$ kg) and a heavy weight ($15$ kg) respectively. Then, we find the $k_{pyl}^{medium}$ as the value at $73\%$ of the $[k_{pyl}^{min}, k_{pyl}^{max}]$ range, according to the optimal trend found and motivated in section~\ref{subsec:BaselineResults}.

\subsection{Hardware}
\begin{figure}[b!] \centering \includegraphics[width=1\linewidth]{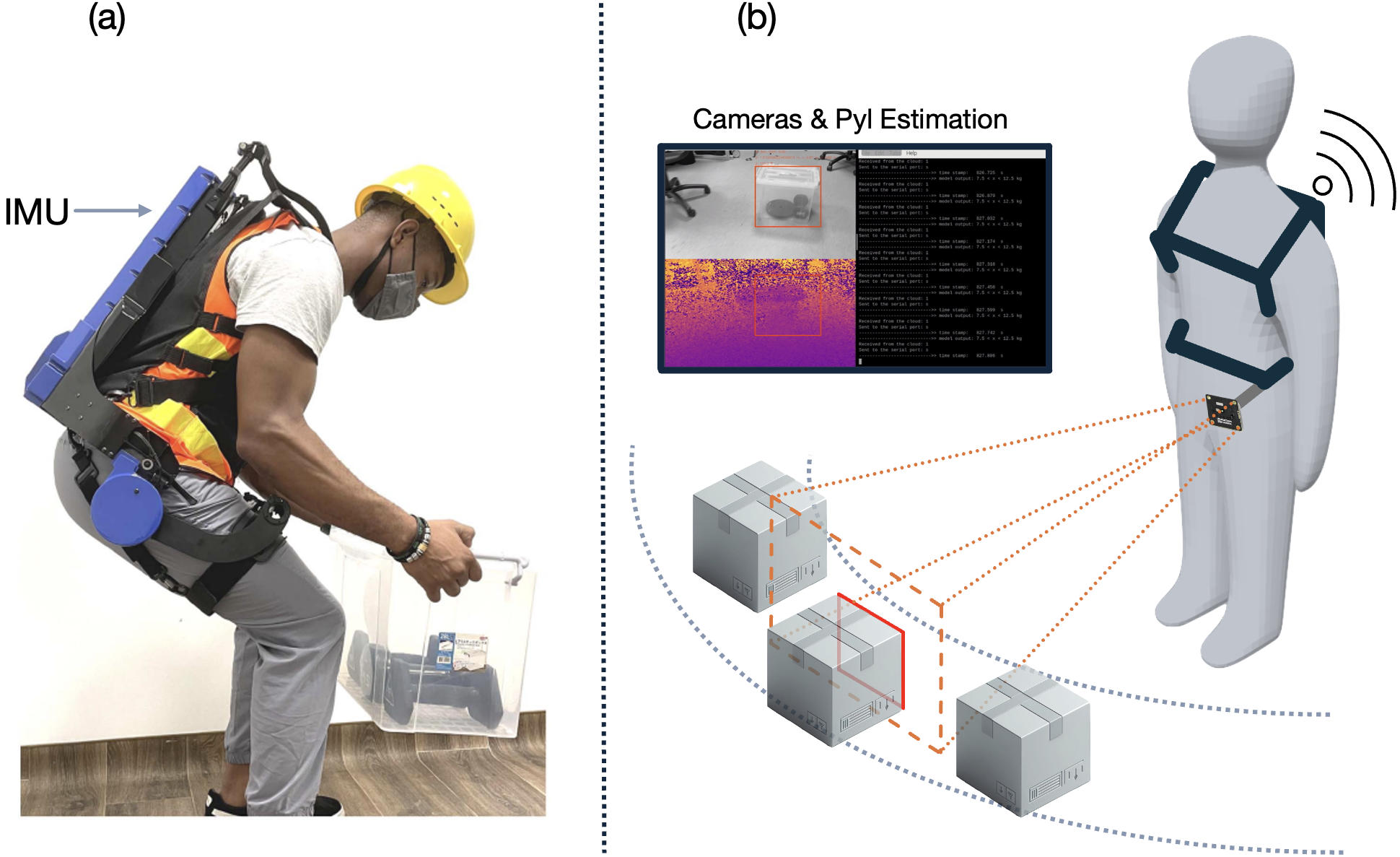} \caption{The back support exoskeleton \cite{Ding} (a) and validation experiments setting (indoor environment) (b).} \label{fig:BackExo&IndoorEnv} \end{figure}
The hardware we used is a single-motor back-support exoskeleton developed at the National University of Singapore \cite{Ding}, shown in Fig.~\ref{fig:BackExo&IndoorEnv} (a). It features a differential series elastic actuator (D-SEA) powered by a $120W$ servo motor (RMD-L-5015) to deliver lifting assistance, with a nominal assistive torque of $30Nm$. The exoskeleton ensures precise force control and active backdrivability, allowing support during lifting while preserving walking autonomy. To enable the computer vision-based adaptive control, we use an RGB camera (12 MP) and an Arducam ToF sensor (240x180 @ 30 fps, $<$2 cm accuracy). We mount both sensors at the right hip (Figure \ref{fig:BackExo&IndoorEnv} (b)), which provides a stable and minimally motion-affected vantage point. Data acquisition is handled by a Raspberry Pi 4B. Aligned with Industry 5.0 principles, we offload the system from the pipeline computational tasks by taking advantage of the cloud. A wireless connection links the exoskeleton to a remote computational station (Apple M3 chip, 8-core CPU, 10-core GPU, 24GB RAM) running the pipeline. After inference, results are returned to the RPI to modulate assistance in real time, as outlined in Figure \ref{fig:Pipeline&PylEstModels}. End-to-end, the computer vision-based adaptive control pipeline achieves an average latency of 100 ms (10 fps) while offloading deep-learning computation to the cloud.

\subsection{Low-Level Control of Back-Support Exoskeleton}\label{subs:low_level_contorl}
As shown in Figure~\ref{fig:LowLevelControl}, the mid-level assistance law computes the desired hip torque $\tau_{\mathrm{cmd}}(\theta,\dot{\theta},\ddot{\theta})$ as the sum of a quasi-static gravity-support term  and a direction-dependent velocity-shaping term:

\begin{equation}
\tau_{\mathrm{cmd}}
= k_\theta\,\theta
- k_v\,\sigma\!\Big(\alpha_e\!\left[\left(\tfrac{\dot{\theta}}{\omega_e}\right)^2
+\left(\tfrac{\ddot{\theta}}{a_e}\right)^2 - 1\right]\Big)\;
\dot{\theta}\,\sqrt{\dot{\theta}^{2}+\varepsilon},
\end{equation}

where $\theta$ is the sagittal-plane trunk/hip tilt angle estimated from an IMU using a real-time complementary filter with coefficient $\alpha=0.98$. The first time derivative, $\dot{\theta}$, is obtained directly from the IMU's gyroscope data without numerical differentiation, while the second time derivative, $\ddot{\theta}$, is computed numerically using a filtered derivative with a time constant of 0.3 s to mitigate high-frequency noise. The term $k_\theta \theta$ provides quasi-static gravity compensation, while $k_v$ scales the magnitude of the velocity-dependent assistance shaping term. We use the sigmoidal activation function $\sigma(\cdot)$ to smoothly gate the dynamic contribution. The quantities $\omega_e$ and $a_e$ are scaling parameters for angular velocity and angular acceleration, respectively, and define, together with $\alpha_e$, an elliptical decision boundary in the $(\dot{\theta},\ddot{\theta})$ plane that distinguishes dynamic motion from near-static conditions. In particular, $\alpha_e$ controls the sharpness of the transition across this boundary. We introduce the small constant $\varepsilon>0$ for numerical regularization around $\dot{\theta}=0$, preventing non-smooth behavior and improving stability near velocity zero-crossings.

Under the adopted sign convention ($\dot{\theta} > 0$ during lowering and $\dot{\theta} < 0$ during lifting), the velocity term reduces assistance during lowering and increases it during lifting, with magnitude increasing approximately with $\dot{\theta}^2$. Compared with the controller in~\cite{Ding}, the present formulation differs in two principal ways: i) it removes hard angle thresholds and replaces them with a smooth motion-dependent gating function, and ii) it explicitly includes angular acceleration in the motion-detection term, improving discrimination between dynamic bending/lifting phases and quasi-static postures.

The resulting torque reference is tracked by the low-level SEA torque controller (PID + disturbance observer), which closes the loop on the measured interaction torque. Payload adaptation is then applied as \begin{equation} \tau_{\mathrm{adj}} = \tau_{\mathrm{cmd}} \cdot k_{\mathrm{pyl}}^{j},\end{equation} where $j \in \{$\textit{light weight}, \textit{medium weight}, \textit{heavy weight}$\}$ denotes the payload class estimated by the high-level computer vision-based adaptive control pipeline, and $k_{\mathrm{pyl}}^{j}$ is the corresponding payload-dependent assistance gain.

We determined the controller gains empirically with the subject wearing the exoskeleton. Participants were asked to repeatedly bend at the hip while the gains were tuned. We initialized $k_\theta$ at a relatively high value and progressively reduced it until the subject reported that the device no longer impeded natural bending. We then selected the nominal value of $k_\theta$ such that the associated torque modulation corresponded to approximately $30\%$ of the gravity-assistance torque at $90^\circ$ of flexion during nominal lifting speeds. We subsequently adjusted the remaining dynamic-term gains to obtain smooth and comfortable assistance during repeated lifting movements.

\begin{figure*}[t!]
    \hspace{-0.5cm}
    \begin{tikzpicture}[auto, >=Latex, node distance=1.2cm and 1.2cm,
        block/.style={draw, rounded corners, minimum height=1.2cm, text width=3cm, align=center, fill=white, thick},
        controller/.style={block, fill=blue!5},
        plant/.style={block, fill=gray!10},
        sum/.style={draw, circle, inner sep=0pt, minimum size=0.6cm, fill=white, thick},
        dashedbox/.style={draw=black!50, dashed, rounded corners, inner sep=10pt}
        ]

        \node (theta) {$\theta$};
        \node[block, right=0.4cm of theta] (gravity) {Quasi-static gravity support\\$k_{\theta}\theta$};

        \node[below=0.8cm of theta] (thetadot) {$\dot{\theta}, \ddot{\theta}$};
        \node[block, right=0.4cm of thetadot] (velocity) {Velocity shaping\\$-k_v\,\sigma(\cdot)\,\dot{\theta}\sqrt{\dot{\theta}^2+\varepsilon}$};

        \node[sum, right=1.0cm of gravity] (sumcmd) {$\Sigma$};
        \node[controller, right=1.0cm of sumcmd] (payload) {Payload adaptation\\$(k_{pyl}^{j})$};
        \node[controller, right=1.0cm of payload] (controller) {SEA torque controller\\PID + DoB};
        \node[plant, right=1.0cm of controller] (actuator) {Motor + SEA spring};
        \node[plant, below=1.2cm of actuator] (sensor) {Torque sensor\\$\tau_{\text{hip}}$};

        \node[above=0.8cm of payload, font=\large\bfseries] (cvac) {Payload classification};

        \begin{scope}[on background layer]
            \node[dashedbox, fit=(gravity) (velocity) (sumcmd), label=above:\textbf{Mid-level assistance law}] (midgroup) {};
        \end{scope}

        \draw[->, thick] (theta) -- (gravity);
        \draw[->, thick] (gravity) -- (sumcmd);
        \draw[->, thick] (thetadot) -- (velocity);
        \draw[->, thick] (velocity) -| (sumcmd);
        \draw[->, thick] (sumcmd) -- node[above] {$\tau_{\mathrm{cmd}}$} (payload);
        \draw[->, thick] (cvac) -- (payload);
        \draw[->, thick] (payload) -- node[above] {$\tau_{\mathrm{adj}}$} (controller);
        \draw[->, thick] (controller) -- node[above] {$i_{\mathrm{cmd}}$} (actuator);
        \draw[->, thick] (actuator) -- node[right] {$\tau_{\text{hip}}$} (sensor);
        \draw[->, thick] (sensor.west) -| node[pos=0.3, below] {$\tau_{\text{hip}}$} (controller.south);

    \end{tikzpicture}
    \caption{Block diagram of the low-level control architecture. The mid-level assistance law combines gravity support and velocity shaping to generate $\tau_{\mathrm{cmd}}$. The computer vision-based adaptive control}-driven gain $k_{pyl}^{j}$ scales the desired torque according to the payload class $j \in \{$\textit{light weight}, \textit{medium weight}, \textit{heavy weight}$\}$ estimated by the high-level pipeline. The SEA torque controller (PID + DoB) compares the spring-deflection torque estimate with $\tau_{\mathrm{adj}}$ to command motor current and regulate the measured hip torque.
    \label{fig:LowLevelControl}
\end{figure*}

\section{Experiments}
\label{sec:Experiments}

\subsubsection*{Demographics \& Ethics}
For both baseline and validation experiments, participants were eligible if they were adults aged $18-65$, with a height of $160-200$ cm, capable of providing informed consent, and in good health to perform weightlifting tasks safely. The experiments took place at the BioRobotics Lab at the National University of Singapore. A total of 24 male healthy subjects (age: $24.29 \pm 2.29$ yo; height: $176.04 \pm 5.30$ cm; weight: $71.98 \pm 8.23$ kg) voluntarily participated in the experiments, and none of the participants reported having a previous low back injury at the time of data collection. Twelve subjects participated in baseline tests, and 12 took part in validation tests, following recommendations for pilot studies to ensure feasibility, precision, and applicability to future research \cite{Julious}. The experiment protocol was approved by the National University of Singapore Institutional Review Board (IRB) Committee with Reference Code LH-20-021, and experiments were conducted in accordance with the Research Compliance Policy on Human Subjects and Biomedical Research. Before the study began, the investigators explained the general purpose of the study and obtained verbal and written informed consent from the participants. 

\subsubsection*{EMG \& Setup}
During all experiments, participants were continuously monitored to ensure safety and avoid accidents. In all experiments, the back-support exoskeleton was utilized in active mode, with muscular activation recorded for the back Erector Spinae Iliocostalis (ESI) and Longissimus (ESL) muscles on both the right and left sides, as well as the leg Biceps Femoris (BF) and Rectus Femoris (RF) muscles. We identified muscle locations following the SENIAM EMG Placement Guidelines \cite{SENIAM}. To standardize data collection, Maximum Voluntary Contraction exercises were performed for each muscle, enabling subsequent signal normalization during analysis. We recorded EMG signals using the Delsys Trigno Wireless System (Delsys Inc., Boston, MA, USA) at a $2150$ Hz sampling frequency. Data analysis adhered to best practices \cite{Konrad}, including the application of a butterworth bandpass filter with low and high cut-off frequencies of $20$ Hz and $450$ Hz, respectively. We rectified signals and calculated the root mean square (RMS) using a $200$ ms sliding time window. 

\subsubsection*{Tasks \& Payloads Conditions}
For baseline tests, where participants performed tasks in fixed positions, we analyzed average and peak activation values over the entire acquisition period. For validation tests involving walking and lifting, we focused the analysis exclusively on lifting cycles.
For all the experiments, we used weights of $5$, $10$, and $15$ kg, also referred to as \textit{light weight}, \textit{medium weight}, and \textit{heavy weight}, based on laboratory availability and safety constraints. We combined gym weights (e.g., dumbbells and plates) of different shapes and types to fill various translucent boxes and obtain the \textit{light weight}, \textit{medium weight}, and \textit{heavy weight} conditions. We continuously varied and combined object and box types to ensure generalization, avoid overfitting to specific box configurations, and mitigate potential effects of differing arm kinematics induced by the boxes, thereby validating our approach across diverse settings. We used translucent boxes to introduce controlled partial observability, forcing the vision pipeline to rely on contextual understanding rather than direct object recognition. This setup maintains consistent lifting biomechanics and reflects realistic industrial scenarios where payloads are manipulated within containers. 

\subsubsection*{Torque Limits}
We used three control strategies (low, medium, and strong assistance levels). Given that the number of assistance-payload combinations increases quadratically with granularity, a finer discretization (e.g., five levels) would have substantially expanded the experimental matrix and prolonged the experimental sessions. For this reason, we limited the number of assistance levels to preserve data quality and avoid muscular and cognitive fatigue. The three-level assistance structure was intentionally selected as an experimental validation step to assess whether the continuous optimal assistance trend identified by our multi-metric analysis could be practically approximated and integrated with a pre-lift payload estimation pipeline. This coarser discretization also ensured reliability, as continuous estimation is sensitive to visual noise and small errors, especially at higher payloads. The 3-class design therefore prioritized stability and safety. The maximum equivalent torque at the hip ranged from 10 Nm to 30 Nm. We conducted preliminary trials with each participant to determine the \textit{low assistance} ($10.80 \pm 0.98$ Nm) and \textit{strong assistance} ($24.00 \pm 1.00$ Nm) torque levels. These values correspond to the mean $\pm$ standard deviation of the torques selected by the participants during the calibration based on subjective perception while lifting 5 kg and 15 kg loads, respectively, within the available torque range. For baseline tests, we defined \textit{medium assistance} as 50\% of the range between \textit{low assistance} and \textit{strong assistance}.
For validation tests, we set medium assistance at 73\% of this range, as informed by baseline results discussed in Section~\ref{subsec:BaselineResults}. 

\subsubsection*{Statistical Methods}
Finally, to formally assess the efficacy of the assistance policies, we analyzed the mean back muscular effort reduction (EMG) and perceived discomfort using Linear Mixed-Effects Models (LMMs). To account for inter-individual variability, we included subject ID as a random effect. We treated Assistance Type (\textit{static assistance}, \textit{adaptive assistance}), Payload (\textit{light weight}, \textit{medium weight}, \textit{heavy weight}), and Muscle Group (L-ESL, R-ESL, L-ESI, R-ESI) as fixed effects. We set statistical significance at $p < 0.05$. The outcomes of these models are detailed in Section \ref{subsubsec:StatisticalAnalysis}.

\subsection{Baseline Experiments}
\label{subsec:BaselineExperiments}
As explained in Section~\ref{sec:ExoOptSpace}, we conducted experiments with 12 subjects to define a two-dimensional optimization space, identifying the appropriate support level based on the payload, considering muscular activity reduction, perceived discomfort, and user preference as performance metrics. Since we classify weights and support levels into three discrete categories (low, medium, and high), there are nine potential combinations to test the exoskeleton's performance. Additionally, three baseline conditions (\textit{light weights}, \textit{medium weight}, and \textit{heavy weight}) without the exoskeleton bring the total to 12 conditions. To further minimize testing time and subject fatigue, we used fractional factorial design \cite{Box2005}, reducing the experiments to eight while maintaining statistical significance. This design samples the boundaries, corners, and midpoint of the optimization space and compares them to the no-exoskeleton baseline. 

For baseline experiments, the subject stands in a fixed position and repetitively lifts a box of varying weights per condition, first without the exoskeleton and then with different support levels. Each lifting cycle is synchronized to 35 beats per minute (bpm) using a metronome. Participants were instructed to perform stoop lifting activities, lowering on every odd beep and lifting on every even beep to standardize movement speed and frequency across subjects. A lifting activity consists of seven cycles, lasting a total of 30 seconds. A minimum rest time of 3 minutes is provided between activities to prevent fatigue or injury. We sampled performance metrics as follows: during preliminary trials, users identified their preferred assistance level (light or strong) for low and heavy weights, yielding samples of $J_{prf}(A_s^i, P_{yl}^i)$. We measured muscular activation in back and leg muscles, as described in Section~\ref{sec:Experiments}, and compared it to non-exoskeleton conditions for the same weight. The percentage reduction in muscular activity provided samples of the EMG-OF $J_{EMG}(A_s^i, P_{yl}^i)$. After each trial with the exoskeleton, subjects completed a questionnaire to evaluate perceived discomfort for each condition as outlined in Section \ref{sec:ExoOptSpace}. The combined ratings from these questions, processed using Equation \ref{eq:Equation3}, provided samples for the perceived discomfort optimization function $J_{dsc}(A_s^i, P_{yl}^i)$.

\subsection{Pipeline Validation Experiments}\label{subsec:ValidationExperiments}
For pipeline validation experiments, we designed a dynamic protocol to evaluate the payload estimation pipeline and the exoskeleton's adaptation capabilities in a more variable environment. The subject starts from a fixed position and walks to three boxes placed in fixed positions ahead, containing $5$, $10$, or $15$ kg each, as shown in Figure~\ref{fig:BackExo&IndoorEnv} (b). To enhance generalization, we shuffle the positions and contents of the boxes before each test. The subject approaches each box, lifts it individually, and returns to the starting position. This process is repeated three times, ensuring each box is lifted three times. We performed this protocol under three conditions: without the exoskeleton, with static light support (preselected based on subject characteristics), and adaptive pipeline control, which, in its “light support” mode, delivers the same torque as the static light support condition. This setup allows us to compare muscular activity across conditions for each lifting cycle. In this protocol, we allowed participants to lift freely at their preferred speed and style. We designed this experimental condition to both challenge the adaptive control in a more dynamic environment and verify whether the optimal trends observed in standardized tests persist under more natural lifting conditions. 

As validation metrics, first of all, we keep track during the experiment of the pipeline classification outputs and ground truth. In this context, because latency is an important concern, we only consider correct classifications happening before the lifting cycle starts. We also measure reductions in muscular activity targeting the same muscles of Section~\ref{subsec:BaselineExperiments}, by comparing the no-exoskeleton condition with the static and adaptive control conditions. We collect user feedback using the same discomfort questionnaire described in Section~\ref{sec:ExoOptSpace}, and we gather user preferences between static and adaptive control. This approach enables us to sample the Validation Functions (VF) $J_{EMG}(A_s^i, P_{yl}^i)$, $J_{dsc}(A_s^i, P_{yl}^i)$, and $J_{prf}(A_s^i, P_{yl}^i)$ for each \textit{static/adaptive assistance} and payload level pairing, creating a final validation space. Using this space, we assess the performance of adaptive control compared to static control. 

\section{Results}
\label{sec:Results}
\begin{figure*}[t!] \centering \includegraphics[width=1\linewidth]{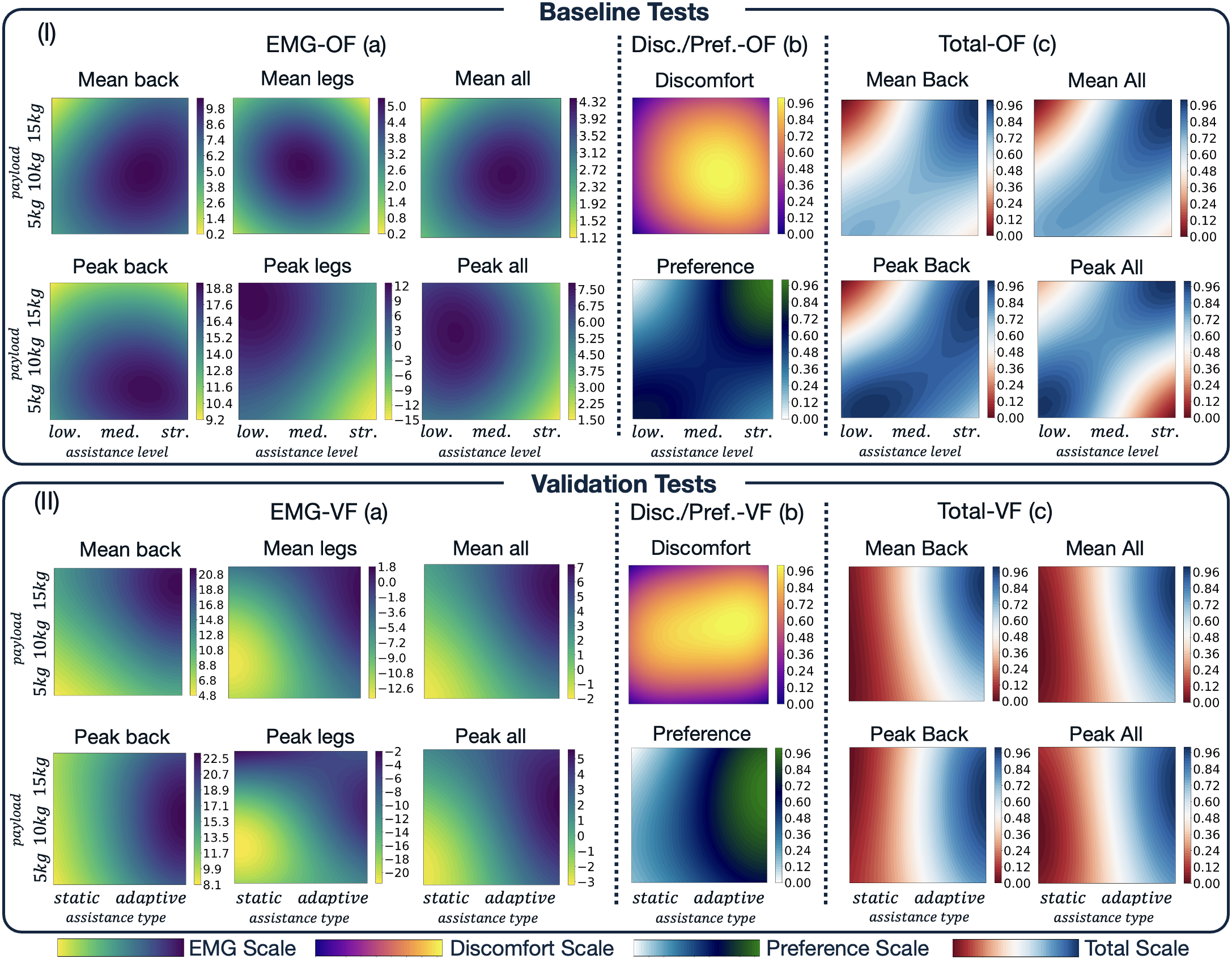} \caption{EMG (a), Discomfort \& Preference (b), and Total (c) optimization functions obtained from Baseline (I) and Validation (II) tests. EMG-based functions, EMG-OF} (Optimization Functions) and EMG-VF (Validation Functions), are reported as mean and peak values for back muscles, leg muscles, and overall muscle contributions. Likewise, Total functions (total-OF and total-VF) are evaluated across all mean and peak combinations for back and overall measures. For EMG functions, the right-hand scale indicates the percentage reduction relative to the no-exoskeleton condition, whereas for the other functions it represents normalized values. 5 kg, 10 kg, and 15 kg correspond to low, medium, and heavy weight conditions. Assistance strategies include \textit{low.}, \textit{med.}, and \textit{str.}, which stand for \textit{low assistance}, \textit{medium assistance}, and \textit{strong assistance} in Baseline tests, and \textit{static/adaptive} (\textit{static} and \textit{adaptive assistance}) in Validation tests.\label{fig:RepresentationFunctions} \end{figure*}
\subsection{Baseline Experiments - Exoskeleton Optimization Space}\label{subsec:BaselineResults}

\subsubsection{EMG-OF}
By analyzing the muscular activation data from the experiments in Section~\ref{subsec:BaselineExperiments}, we derive the EMG-OF $J_{EMG}(A_s, P_{yl})$ shown in Figure~\ref{fig:RepresentationFunctions} (I-a). These functions identify optimal performance areas based solely on muscular activity reduction. For instance, reducing back muscle activity consistently requires high assistance, achieving a $6.32\%$ mean and $16.92\%$ peak reduction for light weights, around $9.62\%$ mean and $15.00\%$ peak reduction for medium weights, and a $6.13\%$ mean and a $9.14\%$ peak reduction for heavy weights. However, high assistance for light weights significantly increases peak activation in leg muscles ($-15.00\%$), which benefit more from medium or low support. The peak EMG-OF for leg muscles indicates that transitioning diagonally from light-weight low-assistance to heavy-weight strong-assistance maintains relatively stable peak leg muscle activation while maximizing mean reduction ($5.00\%$ mean reduction for medium assistance with medium weight). Overall, considering all muscles together, medium to strong assistance is preferable for mean reduction, while medium to low for peak. However, given the adverse effect of strong assistance on peak leg muscle activation, a better approach is to use low assistance for light weights and strong assistance for heavy weights. This strategy balances maximizing back muscle activity reduction while minimizing peak activation in the legs. 

\subsubsection{Discomfort \& Preference-OF}
As outlined in Section~\ref{sec:ExoOptSpace}, we sampled the discomfort function by combining the ratings for each question, adjusting for the negative meaning of discomfort by flipping the ratings of positively worded questions. Using Equation \ref{eq:Equation3}, we obtained discomfort samples for different $(A_s^i, P_{yl}^i)$ pairs. For user preferences, we collected responses on \textit{low assistance/strong assistance} preference for \textit{light weight}/\textit{heavy weight} scenarios. By fitting these samples with Gaussian processes, we derived continuous formulations of $J_{dsc}$ and $J_{prf}$ within the optimization space, as shown in Figure~\ref{fig:RepresentationFunctions} (I-b). The results reveal that perceived discomfort peaks for medium weights with medium-to-high support. Overall, low assistance is preferable to minimize discomfort, regardless of the payload. Conversely, user preferences indicate a tendency to favor high assistance for heavy weights and low assistance for light weights.

\subsubsection{Total-OF}
\begin{figure}[b!] \centering 
\hspace*{-0cm}\includegraphics[width=1\linewidth]{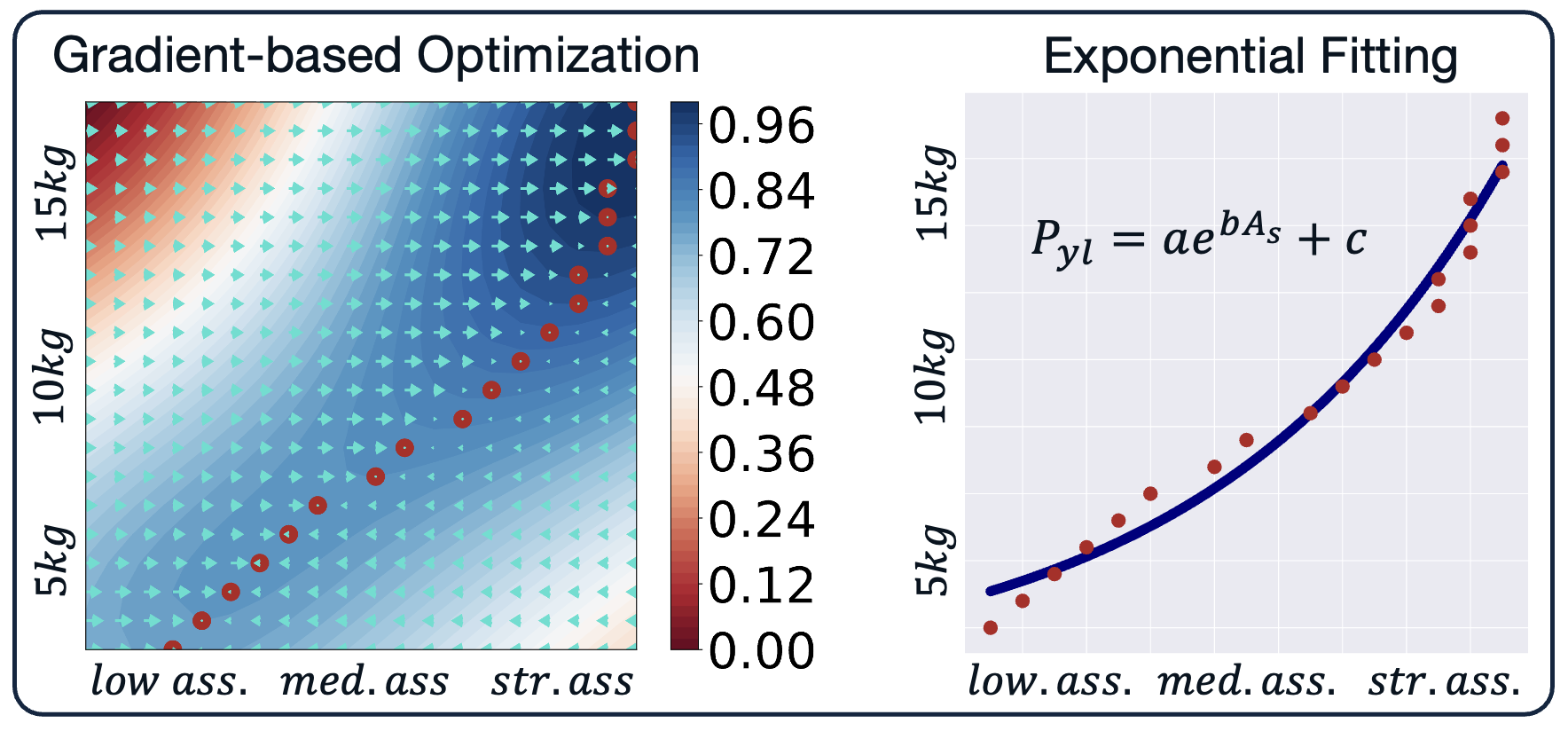} 
\caption{Gradient-based optimization of $J_{tot}$ and fitting process of the optimization points. Arrows show the direction to follow for performance improvement, with red dots indicating the maximum O-RF values.} 
\label{fig:GradientOptimization} 
\end{figure}
It is clear that different OFs' contributions can be considered individually to select the best assistance for a given payload for a specific metric optimization. In our case, to maximize acceptability and effectiveness, we combine the normalized versions of EMG-OF, discomfort-OF, and preference-OF into a total-OF $J_{tot}$, considering EMG contributions such as mean and peak reductions in back and overall muscle activity, as shown in Figure~\ref{fig:RepresentationFunctions} (I-c). Results suggest that to optimize EMG reduction, discomfort, and user preference, assistance should be low for light weights and increase with the payload, following a non-linear trend. Based on this, we focus on optimizing ``Mean All'' activity to minimize muscle activation while avoiding excess leg muscle activation. To determine the optimal assistance for each payload, we compute the partial derivative of $J_{tot}$ with respect to assistance and slice the function along the payload axis. Setting the derivative to zero ($\frac{\partial J_{tot}}{\partial A_s} = f(P_{yl}, A_s) = 0$) for each chosen $P_{yl}$ yields the optimal $A_s^{opt}$ corresponding to each selected $P_{yl}^i$, as shown in Figure~\ref{fig:GradientOptimization}. We subsequently selected an exponential function (see Supplementary Materials for details) to model the relationship between payload and optimal support:
\begin{equation}
    P_{yl} = a \cdot e^{b \cdot A_s^{opt}} + c
\end{equation}
achieving a Root Mean Squared Error $(RMSE)=0.088$ and $R^2=0.977$ with $a = 0.720$, $b = 1.100$, and $c = -1.200$. This function identifies an optimal region in the assistance-payload space, suggesting that if we had a way to continuously estimate payloads, we would always be able to select the best-performing assistance in terms of our defined metrics. Following these findings, instead of the midpoint between low and high assistance levels we select 73\% assistance during validation for medium weights. In the Supplementary Materials, we report the estimated parameters $a$, $b$, and $c$ for the different EMG contributions, together with a detailed justification for the choice of the exponential formulation.
\subsection{MobileNetv2 - DINOv2 comparison}\label{subsec:MobileNetv2&DINOv2}
\begin{figure}[b!] \centering 
\hspace*{-0.20cm}\includegraphics[width=1\linewidth]{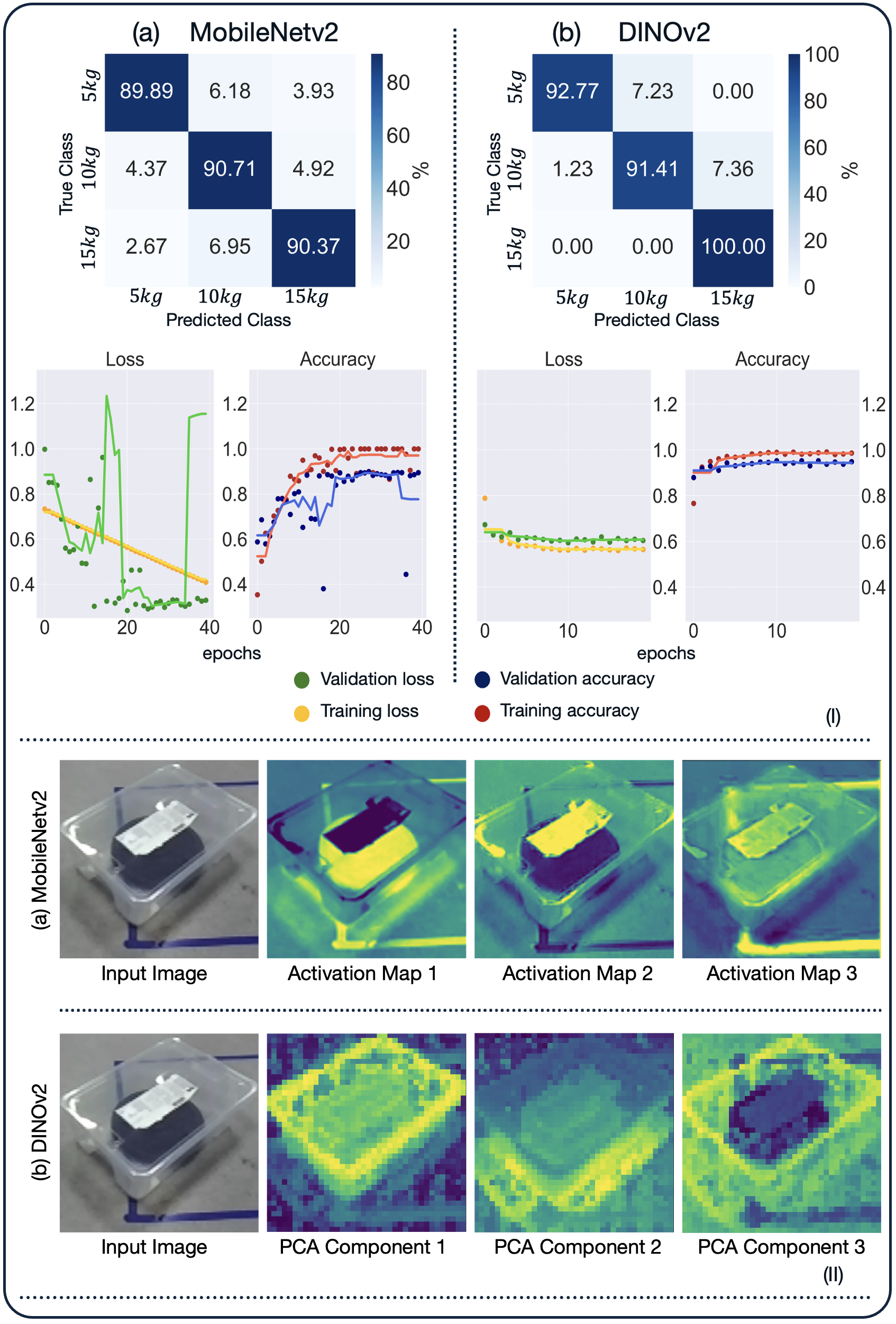} 
\caption{Training history and performance of the payload estimation models (I). Qualitative comparison of feature extraction (II). Where (a) are visualizations of MobileNetv2 convolutional layer activations, highlighting areas of high feature response. And (b) are PCA projection of DINOv2 output features (first three principal components), showing the model's ability to distinguish foreground, background, and depth without explicit supervision.}
\label{fig:Training&Extraction} 
\end{figure}

As discussed in Section~\ref{subsec:PylEst}, we compared two deep learning models, namely MobileNetv2 and DINOv2, primarily used to map input images into meaningful feature spaces. Figure~\ref{fig:Training&Extraction}-I illustrates the offline training and validation performance of the two models. Categorizing box weights based on their contents proved challenging, introducing noise into both training processes, particularly due to dataset variability. Nonetheless, the DINOv2-based model demonstrated more stable, reliable, and less overfitting training compared to MobileNetv2-based one, achieving better performance while trained for a lower number of epochs in terms of higher accuracy improving from $90.32\%$ to $94.72\%$ and average precision from $90.23\%$ to $94.83\%$. We found that these improvements were even more pronounced on other datasets, which we provide in the Supplementary Materials.

To understand these differences, we analyzed the features extracted by both models. For MobileNetv2, we visualized convolutional filters and the activated features during forward passes. For DINOv2, we followed the method in \cite{Oquab}, performing PCA on its output features while only keeping the first 3 components and reshaping them into an image-like format ($width,\:height,\:channels=3$). Figure~\ref{fig:Training&Extraction}-II compares a misclassified image from MobileNetv2 to the same correctly classified image from DINOv2. While both models identified objects within transparent boxes, MobileNetv2 struggled to separate background elements and filter out irrelevant details (e.g., blue tape on the floor and a stack of paper tickets on the box). In contrast, DINOv2 effectively segmented foreground and background, as seen in its feature maps, and demonstrated depth estimation by assigning distinct gradients to pixels based on their distance from the observer. To further support these insights, we conducted a deeper analysis of DINOv2 extracted features across 20 input images, detailed in the Supplementary Material and accompanying code. This analysis included K-means clustering to examine similarities within feature maps across different inputs. Our findings show that DINOv2 consistently assigns similar features to similar objects, such as the background, boxes, box edges, and weights inside the boxes. Each of the three feature maps reveals distinct perception capabilities: PCA component 1 captures background-foreground separation and object detection, PCA component 2 exhibits strong depth perception, and PCA component 3 focuses on detecting weights inside the boxes. We speculate that the final attention block and multilayer perceptron in our architecture integrate these features, enabling more precise payload categorization based on shape, content, and contextual cues. We attribute DINOv2's superior performance to its Transformer-based architecture, which enhances contextual understanding, and its unsupervised learning approach, which produces robust feature representations. This combination allows DINOv2 to outperform state-of-the-art CNNs in detailed object recognition and segmentation tasks. These results drive our choice to use DINOv2 for our real-time experiments.

\subsection{Pipeline Validation}\label{subsec:ValidationResults}
\begin{figure}[b!] \centering 
\hspace*{-0.5cm}\includegraphics[width=1\linewidth]{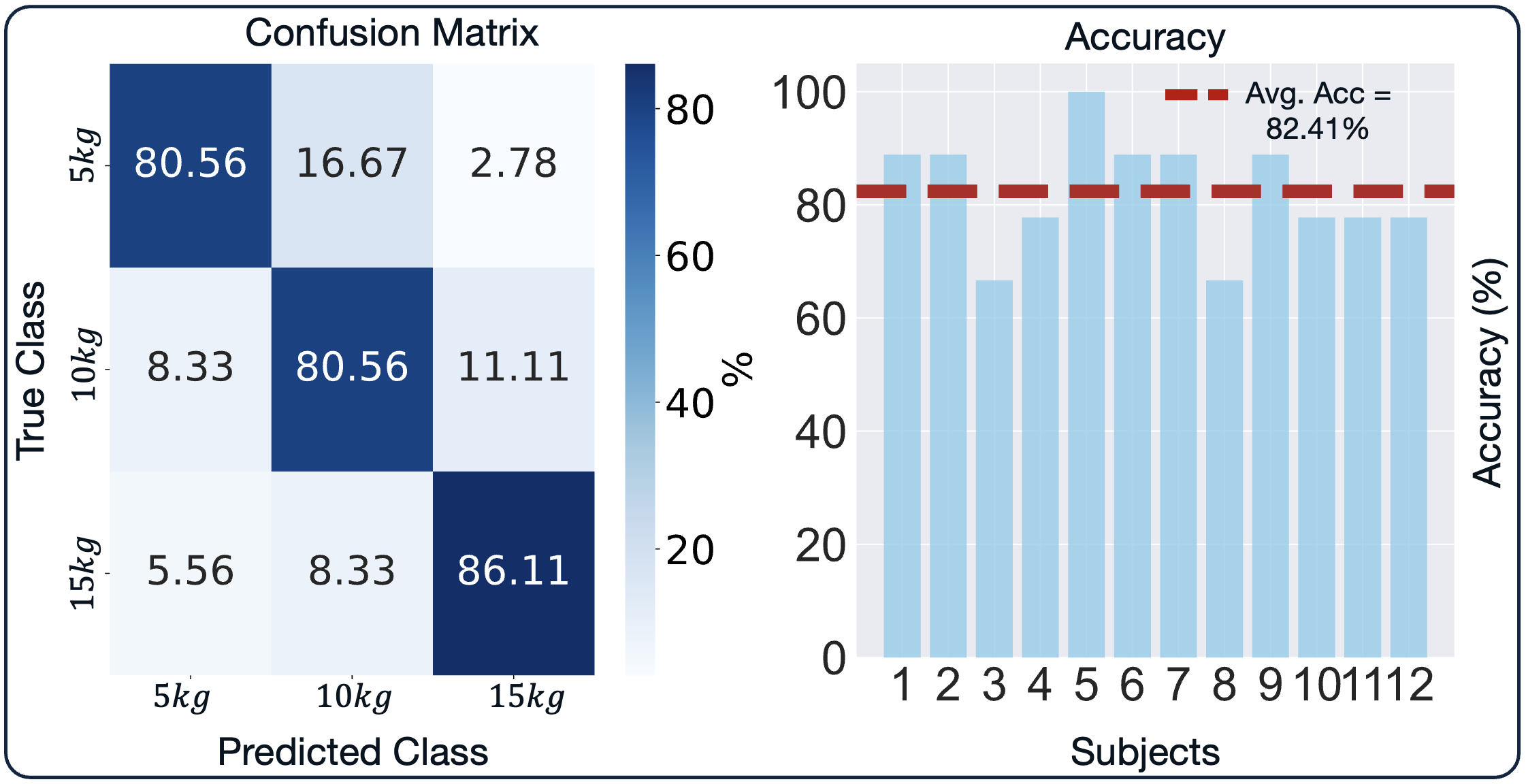} 
\caption{Pipeline classification performances in terms of confusion matrix and subject-specific average accuracies.} 
\label{fig:PipelineClassificationPerformance} 
\end{figure}

\subsubsection*{Muscular activation results}
As outlined in Section~\ref{subsec:ValidationExperiments}, during experiments we measure muscular activity to be compared with no-exoskeleton condition, we track output classifications of the pipeline, and ensure that each classification is correct only if the correct output of the classification is provided before the lifting cycle starts. We also collect human feedback on discomfort and preference similarly to what was done for baseline experiments to build validation functions. As shown in Figure~\ref{fig:RepresentationFunctions} (II-a), regarding back muscle activation, we observed unexpected changes under the light weight condition, even though the low assistance selected by the pipeline should be the same as the one provided during the static assistance test. We registered improvements in the case of adaptive control ($4.90\rightarrow11.07\%$ for mean and $8.35\rightarrow19.22\%$ for peak activation), which we attribute to reduced overall effort during the test, and some light weight misclassification leading to stronger support. We registered a strong improvement for the medium and heavy weight conditions, where adapting the control to the handled weight strongly improves muscular activation reduction compared to static control condition ($9.13\rightarrow18.89\%$ for mean and $9.62\rightarrow23.05\%$ for peak activation for medium weight and $12.15\rightarrow20.68\%$ for mean and $9.99\rightarrow19.05\%$ for peak activation for heavy weight). We obtain similar results in terms of gradients when combining all of the muscles' contributions together. We achieve instead a particularly interesting result when considering leg muscles, which also experience a decrease in peak ($-12.35\rightarrow1.08\%$ for medium weight) and mean ($-19.33\rightarrow-2.65\%$ for medium weight) muscle activation when using an adaptive control compared to a static one during lifting. Overall, adaptive control shows strong improvements in EMG reduction across all the muscles. 

\subsubsection*{Discomfort \& Preference results}
In alignment with baseline tests, discomfort-VF suggests an increase in perceived discomfort going towards adaptive control, especially marked for medium and heavy weights, while remaining almost unchanged when handling a light weight. This is a result aligned with baseline results, where stronger support led to increased perceived discomfort. preference-VF also aligns with results obtained from the baseline experiments: despite the discomfort slightly increasing, subjects still prefer higher assistance for heavier weights, leading to preferring adaptive control compared to static one. Finally, all the total-VF display gradients pointing towards adaptive control for each of the payload conditions. In conclusion, 83.33\% of the subjects agreed in feeling safe letting the exoskeleton autonomously decide which control to select. 

\subsubsection*{Pipeline performance}
Concerning the computer vision-based adaptive control performances, using video recordings of the experiments, we evaluated the accuracy of the object detection and selection layer described in Section \ref{subsec:HL_Control_Strategy_ForBE}. The method performed robustly, achieving 98.13\% accuracy in real-time trials. While Figure~\ref{fig:PipelineClassificationPerformance} shows the final validation performances of the whole computer vision-based adaptive control pipeline in terms of confusion matrix and accuracies per each subject test and average final value. During real-time tests, the pipeline achieves 85.30\%, 76.30\% and 86.10\% precisions for the light, medium, and heavy weight classes respectively, and an overall 82.41\% accuracy. Overall, the pipeline shows balance between overestimations and underestimations of the weights, suggesting that no particular bias is present. 

\subsection{Statistical Analysis of Validation Results}\label{subsubsec:StatisticalAnalysis}
\begin{figure}[b!] \centering 
\hspace*{0cm}\includegraphics[width=1\linewidth]{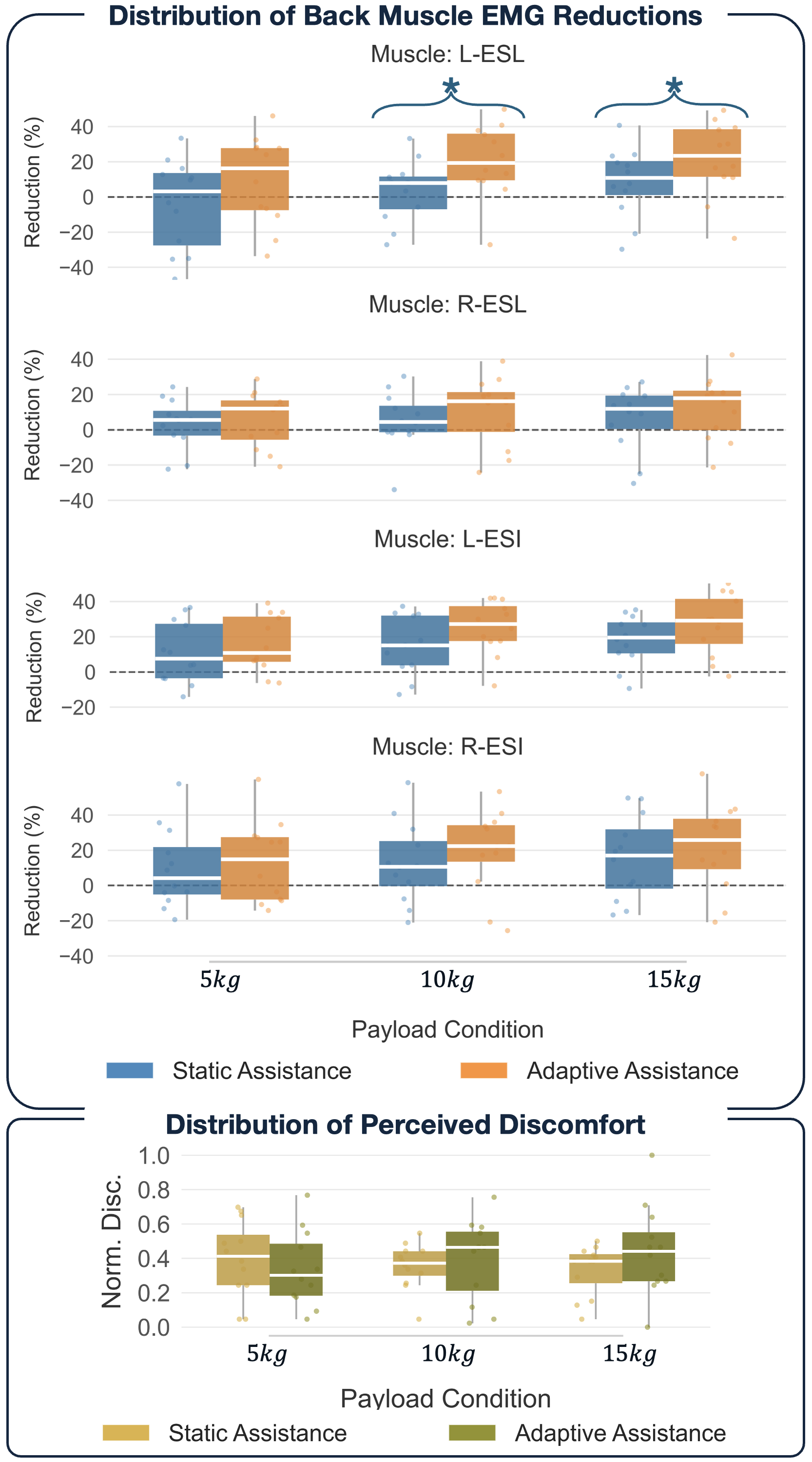} 
\caption{Distributions of back muscle electromyography reduction for static assistance and adaptive assistance, where L/R-ESL = Left/Right Erector Spinae Longissimus and L/R-ESI = Left/Right Erector Spinae Iliocostalis. Boxes span the interquartile range (25th to 75th percentiles), white lines indicate the median, and whiskers extend to 1.5 times the interquartile range. Raw data points are overlaid to illustrate population variance. Statistically significant differences ($p < 0.05$) from post-hoc analysis are explicitly marked.}
\label{fig:BackEMGDistribution} 
\end{figure}

To properly account for the high inter-subject variability inherent in biomechanical data, we performed a Linear Mixed-Effects Model analysis using the \textit{statsmodels} library in Python. The model treated assistance type, payload, and muscle group as fixed effects, while we included participant as a random group effect to isolate within-subject variation. We evaluated the assumption of normality for the LMM residuals through visual inspection of Quantile-Quantile (Q-Q) plots and confirmed statistically via the Shapiro-Wilk test (EMG model: $W = 0.9915$, $p = 0.0959$; discomfort model: $W = 0.9804$, $p = 0.3213$). The model revealed a highly significant main effect of assistance type ($p < 0.001$). Specifically, the model coefficient demonstrated that adaptive assistance provided an 8.15\% greater overall reduction in back muscle activity compared to static assistance (Cohen's $d=0.405$). We also observed a significant main effect of payload ($p < 0.001$), with higher offloading achieved in the heavy weight condition relative to the light weight condition, as well as a significant main effect of muscle group ($p < 0.001$), where inner iliocostalis muscles exhibited greater reductions than outer longissimus muscles.

Figure \ref{fig:BackEMGDistribution} illustrates the raw, unadjusted population data distributions. While the Linear Mixed-Effects Model confirms the overarching effectiveness of adaptive assistance across the dataset, post-hoc paired t-tests with Benjamini-Hochberg false discovery rate corrections applied to specific conditions highlight the underlying population variance. Significant localized differences between \textit{adaptive} and \textit{static assistance} were observed specifically for the left Erector Spinae Longissimus muscle under medium weight ($p = 0.011$) and heavy weight ($p = 0.011$) conditions.

In contrast, a Linear Mixed-Effects Model applied to the normalized perceived discomfort questionnaire data (scaled from 0 to 1) showed no significant main effects for either assistance type ($p = 0.454$) or payload ($p = 0.679$). Overall, these statistical models demonstrate that adaptive assistance provides a measurable improvement in physical offloading across varying payloads without negatively impacting perceived user discomfort.

\section{Discussion}
\label{sec:Discussion}
\subsection{Exoskeleton Optimization Space and Contributions}
Results in Section~\ref{subsec:BaselineResults} and Figure~\ref{fig:RepresentationFunctions} (I) indicate that different support modulation strategies can optimize distinct performance metrics. For instance, maintaining high support effectively reduces both peak and mean back muscle activation. However, this approach increases peak activation in leg muscles, particularly during light-weight lifting. While not extensively documented, some research has indicated a comparable pattern, wherein the energy imparted by a non-grounded back-support exoskeleton seems to distribute throughout the body, possibly reducing back muscle engagement but also, under specific conditions, slightly altering leg loading \cite{Plad}, \cite{Luger2023}. Given the negative impact on peak leg muscle activation when using strong support for light weight, an optimal approach for minimizing overall muscular effort would involve high support for reducing mean activation and lower support to mitigate peak activation. Alternatively, considering back and leg muscle activity optimization separately, the best strategy would involve increasing support progressively with weight, from low to strong, ensuring a stable reduction in leg muscle activation. Regarding discomfort, results show that increasing support, regardless of payload, leads to higher perceived discomfort. This finding is consistent with our experience and participants' feedback, which suggest that greater support exacerbates joint misalignment, increases human-exoskeleton contact forces, and restricts natural movement, all contributing to discomfort. Comments gathered during experiments highlight that strong assistance made bending down more challenging, worsened joint misalignment, and caused localized pressure, particularly at anchoring points such as the thighs and shoulders. Nonetheless, user preference indicates a tendency to favor strong support for heavy loads and low support for lighter ones. This likely stems from a perceived benefit: for short-duration tasks like single-cycle lifting, the exoskeleton’s back muscle relief outweighs the discomfort, whereas for light loads, users do not find the additional support necessary, so they prefer to prioritize discomfort minimization. Once again, the preference function suggests that adaptive support is the most effective strategy. In the case of total-OF, the support modulation trend is evident across all EMG-OF contributions, highlighting that maintaining a static support level is suboptimal for maximizing EMG reduction, minimizing discomfort, and aligning with user preferences. Instead, an adaptive control strategy adjusting support from light to strong based on weight would yield the best overall performance. Although it is intuitive that heavier payloads require greater assistance, the quantitative, multi-metric landscape that determines the optimal level of assistance balancing multiple metrics has not been characterized in prior work. In this study, we provide the first systematic and empirical analysis of this landscape and show, through the results in Figure~\ref{fig:GradientOptimization}, that it follows a non-linear, subject-independent exponential trend that cannot be inferred from intuition alone. To the best of our knowledge, this is the first study to jointly consider multiple metrics in order to derive an optimized assistance profile for back exoskeletons.

\subsection{Payload Estimation Models and Importance of Computer Vision-Based Techniques to Beat Latency}
Our results in Section~\ref{subsec:MobileNetv2&DINOv2} confirm that payload estimation from transparent boxes remains a challenging task. Despite achieving high accuracy ($90.32-94.72$\%), factors such as varying viewpoints, light reflections, and occlusions caused by box shapes can still lead to misclassification. The MobileNetv2-based model exhibits unstable learning curves, raising concerns about its real-time reliability, even though the final epoch achieves high validation accuracy. Conversely, state-of-the-art vision transformer architectures like DINOv2 demonstrate more stable training (Figure~\ref{fig:Training&Extraction}-I) and superior overall performance. While it is less pronounced in DINOv2, both models exhibit some degree of overfitting. However, since our objective is not to develop a zero-shot payload estimation model and industrial environments are typically standardized, training on a dataset collected within the same setting minimizes the risk of encountering vastly different objects, making this overfitting effect acceptable. Feature learning from raw images has been central to deep learning-based computer vision, particularly in CNN applications \cite{Voulodimos}, \cite{Beaglehole}. However, vision transformer architectures, especially those trained in an unsupervised manner \cite{Caron, Darcet2024}, have shown superior general feature extraction capabilities, outperforming CNNs in various applications. Our findings further support this, as feature map analysis confirms that DINOv2 extracts higher-quality features, enabling more advanced image detection, segmentation, and depth perception capabilities that standard CNNs, even optimized ones, struggle to achieve. Figure~\ref{fig:Training&Extraction}-II illustrates how these advantages allow the DINOv2-based model to distinguish foreground from background, filter out irrelevant elements such as floor markings and labels, and demonstrate depth perception crucial for volume and shape estimation. These advanced capabilities result in correct classifications for DINOv2, while MobileNetv2 more frequently misclassifies objects. Real-time evaluations confirm strong performance. While our object detection and selection layer achieved 98.13\% accuracy, we acknowledge that it may still benefit from additional hyperparameter tuning in more challenging environments. As shown in Figure~\ref{fig:PipelineClassificationPerformance}, the end-to-end system achieves a mean accuracy of $82.41$\%, with subject-specific accuracies closely distributed around this mean, indicating that our approach effectively mitigates the subject-specific variability found in other payload estimation methods \cite{Xiao}. Accuracy drops for subjects 3 and 8 are primarily due to communication conflicts between the cloud computing platform and the embedded Raspberry Pi, arising from interference between the embedded computer and the EMG wireless sensor system, both operating at the same frequency. While this highlights a limitation in our pipeline, it also suggests that improved system design could further enhance stability and accuracy. Additionally, the confusion matrix shows that misclassifications are evenly distributed around the diagonal, indicating a well-balanced model without inherent bias. All correctly classified transitions occurred before the lifting cycle began, ensuring prompt detection and support modulation. To the best of our knowledge, this is the only technique available to date that enables payload estimation and support modulation in advance, maximizing exoskeleton effectiveness and benefits. Our results in Figure~\ref{fig:RepresentationFunctions} (II), particularly in terms of EMG-VF, strongly support this claim. 

\subsection{The Importance of Adaptive vs Static Control}

\subsubsection*{Muscular activation results}
Results in Section~\ref{subsec:ValidationResults} indicate that, except when optimizing solely for discomfort, adaptive assistance is consistently preferred. Figure~\ref{fig:RepresentationFunctions} (II) shows that adaptive control enhances EMG reduction across all contributions, even in cases where support levels should theoretically be the same, such as for light weights, where both static and adaptive assistance provide low support. An interesting finding is the increase in peak and mean leg muscle activation observed during experiments, particularly for the light-weight static assistance condition, which slightly deviates from baseline results. In addition to the pipeline's non-perfect classifications, we attribute this to differences in test protocols: while baseline tests involved controlled, standardized stoop lifting, validation tests allowed participants to move freely, lifting at their preferred speed and with their preferred style. As a result, some subjects performed a mixed stoop-squat lifting motion, leading to increased leg muscle activation. Video analysis and EMG data confirm this effect, particularly for subjects 2 and 8. This effect suggests that adaptive control allows subjects to lift weights more comfortably without relying on leg muscles, leading to greater muscle activity reduction in both the back and legs. For payload increasing static support does not substantially enhance muscle activity reduction, whereas adaptive support does, as expected, by dynamically adjusting support based on payload. This trend is evident across all contributions. 

\subsubsection*{Subjective feedback}
Regarding discomfort, the VF results align with expectations, showing an average increase under adaptive control, even though our statistical analysis does not show any evidence of overall higher discomfort. Nevertheless, this average trend supports the baseline findings that higher support levels lead to greater discomfort, as adaptive control adjusts assistance based on payload, while static support maintains low assistance regardless of weight. Similarly, the preference-VF indicates that most subjects favored adaptive control, as it provided greater support for heavier loads while minimizing discomfort for lighter ones by adapting to low assistance. When considering all factors together, the total-VF results reveal a clear trend favoring adaptive control for performance optimization. As noted in the experiments section, we intentionally adopted distinct protocols to test whether optimal performance trends observed in standardized conditions persist in more dynamic environments. The results confirm this, as adaptive control remained preferable to the static strategy even under more variable lifting conditions.

\subsection{Statistical Significance of Validation Results}

The robustness of these findings is supported by explicitly modeling inter-subject variability. Human physiological responses and subjective perceptions are inherently variable, and this natural variance can obscure true intervention effects when analyzing raw population data. By incorporating random intercepts for each participant, the Linear Mixed-Effects Model analysis (detailed in Section \ref{subsubsec:StatisticalAnalysis}) filters out these baseline differences. This is reflected in the large between-subject variance observed in the electromyography model ($\sigma^2_{group} \approx 192.1$), highlighting why this approach was essential. The significant back muscle activity reductions associated with the adaptive assistance policy ($p < 0.001$) therefore represent a consistent, overarching effect across participants rather than cohort-specific noise. Simultaneously, the absence of significant differences in discomfort ($p = 0.454$) indicates that the adaptive control does not negatively impact user perception, despite its dynamic nature. This is particularly relevant in wearable robotics, where time-varying assistance can introduce instability. Finally, the preference function in Figure \ref{fig:RepresentationFunctions} (II-b) further supports these findings, indicating a user preference for adaptive over static assistance.

The statistical advantage of the adaptive strategy over the static baseline highlights an important advancement in exoskeleton control design. While static assistance profiles provide a consistent baseline reduction in muscular effort, similar to the fixed nominal torque strategies successfully validated on this hardware platform in prior work~\cite{Ding}, our results indicate that such static assistance does not fully exploit the potential for performance improvement under varying load conditions. By explicitly incorporating pre-lift contextual information, the dynamic modulation of adaptive assistance offers a distinct advantage. The proposed adaptive strategy achieves statistically significant improvements over fixed-assistance baselines for heavier loads, while maintaining comparable levels of leg muscle activation and user discomfort for lighter tasks. These statistically supported findings suggest that context-aware modulation provides a more effective balance between biomechanical effectiveness and user comfort in variable operational scenarios.

Eventually, our findings support the claim that adaptive control is important for performance optimization, suggesting a preference for active exoskeletons over passive ones to achieve this level of adaptability.

\subsection{Limitations \& Future Work}
Our pipeline shows promising results in real-time detection and validation, demonstrating that adaptive control can significantly improve performance. However, it has limitations that may hinder adoption in industrial settings. A key limitation of this study is that payload estimation was tested primarily with transparent boxes, which simplify detection and inference. Estimating weights for opaque or irregularly shaped objects, common in industrial settings, remains challenging. While opaque boxes with attached payload information (e.g., QR codes or labels) can be handled by fine-tuning DINOv2, estimating weights for irregular or opaque objects without explicit cues remains challenging and may benefit from additional sensing, such as sensor fusion or EMG/FSM data. Human motor control achieves robust performance through a combination of predictive (visual) and reactive (tactile) mechanisms. In this work, we isolate and validate the predictive visual component, while future extensions could integrate it with real-time haptic feedback to achieve high-performance, dual-modality exoskeleton assistance, mirroring the reliability of the human sensorimotor system. Furthermore, incorporating dual heterogeneous sensors and sensor-fusion strategies would provide hardware-level redundancy and fault-tolerant monitoring, which are necessary to meet safety requirements for real-world deployment. Additionally, our current pipeline runs on the cloud, which is suitable for the experimental validation of our framework and aligns with the Industry 5.0 paradigm. However, it also introduces certain limitations, such as reduced system autonomy and potential real-time issues caused by communication delays. Future work will therefore focus on software optimization and the integration of edge-computing capabilities to ensure reliable, low-latency performance even in environments with unstable network connectivity. Ultimately, our goal is to enable fully embedded execution without increasing the bulk or weight of the exoskeleton. Another limitation is that our method was tested only in a controlled lab with young subjects and a three-level assistance scheme. While suitable for the experimental validation, future research efforts should be put toward continuous assistance modulation, improved payload estimation, and testing with more diverse objects. Advancing beyond the proof of concept will require evaluations in real industrial environments with varied user profiles to assess scalability, robustness, and latency performance.

\section{Conclusion}
\label{sec:Conclusion}
In this research, we addressed two key questions: identifying optimization spaces for back-support exoskeletons to determine optimal operating areas and exploring control adaptation strategies to enhance performance through exoskeleton context awareness. We established a two-dimensional optimization space using Optimization and Validation Functions based on EMG reduction, perceived discomfort, and user preference. Experiments with 12 subjects revealed optimal operating regions, demonstrating that support modulation is essential for performance optimization. Building on these insights, we developed a computer vision-based adaptive control pipeline. Unlike previous approaches, our method is subject-independent, does not increase exoskeleton bulk, and minimizes latency by estimating payloads before lifting. Integrated into the exoskeleton control and validated with 12 subjects, the pipeline demonstrated strong real-time accuracy and effectiveness across performance metrics. Our findings highlight the importance of adaptive control in optimizing exoskeleton performance and suggest that active exoskeletons could be preferred over passive/static ones to achieve the necessary adaptability. These results pave the way for further research in back exoskeleton optimization, context awareness, and high-level control adaptation.

\section*{Data \& Code Availability Statement}
This article has downloadable data available at
\href{https://zenodo.org/records/15125115}{10.5281/zenodo.15125114}, provided by the authors. Further Supplementary Material on pipeline development details and shared code can be found at the following GitHub repository: \href{https://github.com/Andredp9/Computer-Vision-Based-Adaptive-Control-for-Exoskeleton-Performance-Optimization.git}{GitHubRepository}.

\vspace{-0.2cm}
\bibliographystyle{IEEEtran}
\bibliography{bibliography.bib}

\end{document}